\documentclass[11pt,a4paper]{article}
\usepackage[hyperref]{acl2020}
\usepackage{times}
\usepackage{latexsym}
\usepackage{graphicx}
\usepackage{subcaption}

\usepackage{microtype}

\aclfinalcopy 

\title{NUBIA: NeUral Based Interchangeability Assessor for Text Generation}

\author{
  Hassan Kane \thanks{Equal contribution. } \\
  WL Research \\
  \texttt{hassanmohamed@alum.mit.edu}
  \\\And
   Muhammed Yusuf Kocyigit \footnotemark[1]\\
   WL Research \\
  Bogazici University \\
  \texttt{yusuf.kocyigit@boun.edu.tr}
  \\\And
  Ali Abdalla\\
   WL Research \\
   \texttt{aabdalla@alum.mit.edu} \\
  \\\AND
   Pelkins Ajanoh \\
   WL Research \\
  \texttt{pelkins@alum.mit.edu} \\
  \\\And
  Mohamed Coulibali \\
  WL Research\\
  Laval University \\
  \texttt{mohamed-konoufo.coulibaly.1@ulaval.ca} \\
}
\date{}

\begin{document}
\maketitle
\begin{abstract}
We present NUBIA, a methodology to build automatic evaluation metrics for text generation using only machine learning models as core components. A typical NUBIA model is composed of three modules: a neural feature extractor, an aggregator and a calibrator. We demonstrate an implementation of NUBIA which outperforms metrics currently used to evaluate machine translation, summaries and slightly exceeds/matches state of the art metrics on correlation with human judgement on the WMT segment-level Direct Assessment task, sentence-level ranking and image captioning evaluation. The model implemented is modular, explainable and set to continuously improve over time.
\end{abstract}

\section{Introduction}

Evaluation metrics play a central role in the machine learning community. They direct research efforts and define the state of the art models. In many text generation tasks, especially in machine translation and summarization, the two most common metrics used for evaluating similarity between candidate and reference texts are BLEU (Bilingual Evaluation Understudy)\citep{papineni2002bleu} and ROUGE (Recall-Oriented Understudy for Gisting Evaluation)\citep{lin2004rouge}. Both approaches rely on counting the matching n-grams in the candidate text to n-grams in the reference text. The former is precision focused while the latter is recall focused. 

These metrics have posed serious limitations and have already been criticized by the academic community \citep{reiter2018structured} \citep{callison2006re} \citep{sulem2018bleu} \citep{novikova2017we}. In this work, we present a methodology to build text generation evaluation metrics using deep learning models as core components.  

An implementation of this methodolgy is then presented and tested in the domains of machine translation and image captioning. For assessing the metric in the machine translation domain, we use the WMT 2017 segment-level Direct Assessment task and show that our method outperforms all current metrics in terms of absolute pearson correlation with human judgement of machine translation quality. It is then shown to closely match state-of-the art metrics in terms of ranked correlation of machine translation for the WMT 2018 and 2019 dataset. We conduct further experiments showing that it also outperforms existing metrics used to assess image captioning quality. Beyond the promise of this methodology in terms of its correlation with human judgment, NUBIA metrics are explainable, can be constructed with any base architecture, and expected to improve continuously with future advances in semantic similarity, linguistic inference and language modeling. 

\section{Related Work}
\subsection{BLEU, ROUGE and n-gram matching approaches}

BLEU\citep{papineni2002bleu} and ROUGE \citep{lin2004rouge} have been used as the main evaluation methods in a variety of NLP tasks for almost two decades. BLEU is shown to better correlate with human judgment when the hypothesis texts are bad as we can see in figure \ref{visuals}(c) and correlate weakly when the hypothesis texts are better. CIDEr  is an image captioning metric that computes cosine similarity between tf–idf weighted n-grams \citep{cider2014}. These methods tend to perform better as the number of reference sentences grow. With recent advances in the quality of text generation, these metrics have been failing to correctly evaluate the performance of models and are creating a bottleneck in the progress of generative models. While the general acceptance of these methods depend on many factors including their simplicity and intuitive interpretability, the core claim that they highly correlate with human judgement \citep{papineni2002bleu} does not hold up anymore.  

The shortcomings of these methods have been widely criticised and studied. \citet{reiter2018structured}, in his structured review of BLEU, finds a low correlation between BLEU and human judgment. \citet{callison2006re} examine BLEU in the context of machine translation and find that BLEU neither correlates with human judgment on adequacy (whether the hypothesis sentence adequately captures the meaning of the reference sentence) nor on fluency(the quality of language in the hypothesis sentence). \citet{sulem2018bleu} examine BLEU -- in the context of text simplification -- on grammaticality, meaning preservation and simplicity. They report a very low, and, in some cases, negative correlation with human judgment.

Considering these results, it is a natural step to pursue new avenues for text generation evaluation and, with the advent of deep learning, using neural networks for this task is a promising step forward. Deep learning models can go beyond n-gram matching and encode higher order features important for assessing machine translation quality such as semantic relatedness of the sentences, logical entailment and grammaticality of the candidate sentence.

\subsection{Transformers, BERT and GPT}

Language modeling has become an important NLP technique, thanks its ability to be applied to various NLP tasks as explained in \citep{radford2019language}. There are two leading architectures for language modeling: Recurrent Neural Networks (RNNs)\citep{mikolov2010recurrent} and Transformers \citep{vaswani2017attention}. RNNs handle the input tokens, words or characters, one by one through time and learn the relationship between them, whereas, transformers receive a segment of tokens and learn the dependencies between them using an attention mechanism.

The recent success of transformers as multitask learners\citep{radford2019language} motivated us to adapt them for the task of neural language evaluation. This is crucial because what stood as an obstacle before neural language was the power to generalize well to different datasets and tasks. Now with models like GPT-2\citep{radford2019language} and RoBERTa\citep{liu2019roberta}, trained on huge amounts of data we can start trusting their ability to generalize across domains. As of now, machine summarization, translation and image captioning all use different metrics to compare reference sentences with candidate sentences. Transformers-based models offer the promise to unify evaluation across these text generation tasks.

\subsection{Model-based metrics}

While BLEU and ROUGE are defined in a discrete space of word tokens, some new evaluation metrics are utilizing neural networks and are defined in the continuous space of word vectors. BERTscore \citep{zhang2019bertscore} uses word embeddings and cosine similarity to create a score array and uses greedy matching to maximize the similarity score. Sentence Mover’s Similarity \citep{clark2019sentence} uses the mover similarity, Wasserstein distance, between sentence embedding generated from averaging the word embeddings in a sentence. YiSi \citep{lo2019yisi} also defines a distance metric among reference and hypothesis sentences based on multilingual BERT embeddings and word frequency weightings. SPICE\citep{anderson2016spice} is an image captioning metric that creates a parse tree from the reference caption, candidate caption to create a scene graph and compute a score based on the overlapping relationships.

 These methods report stronger correlations with human judgment and better results when compared with BLEU and ROUGE. While they are using word embeddings \citep{mikolov2013distributed} to transfer their sentence in a continuous space, they use hand-crafted mathematical functions to evaluate similarity in that space. In NUBIA, rather than defining a mathematical formula, we train a neural network to learn it using human judgement on thousands of sentence pairs as signal. 

One other evaluator that uses machine learning is BLEND \citep{ma2017blend} which uses an SVM to combine different existing evaluation metrics. 

Another proposed evaluation method is RUSE \citep{shimanaka2018ruse}. This method embeds both sentences separately and pools them to a given size. After, the method uses a pre-trained MLP to predict on different tasks. This quality estimator metric is then proposed to be used in language evaluation. 

Our proposed methodology, is also a learned metric. We are proposing to use different pre-trained transformers to extract features on reference and hypothesis sentences and later train an aggregating module to predict the final quality score. 

\subsection{GLUE Benchmark}

The GLUE Benchmark is a collection of tools for evaluating and analyzing the performance of models across a diverse range of existing NLU tasks \citep{wang2018glue}. The recent introduction of this benchmark has catalyzed the development of architectures scoring well on a wide variety of tasks and encouraged the NLP community to move away from specialized models doing well on a single task to models performing well across diverse tasks.

The variety of tasks introduced in the GLUE Benchmark are linguistic acceptability, sentiment analysis, semantic similarity, question answering, logical inference and reading comprehension. To be assessed according to that benchmark, models such as Transformers are usually pre-trained on a large corpus in an unsupervised manner and fine-tuned on a dataset used for the specific task of the benchmark.

\section{Feature-based Neural Language Evaluator}
Our method has three modules: a neural feature extractor, an aggregator and a calibrator. The feature extractor tested in this paper consists of different transformer based architectures fine-tuned on relevant tasks of language evaluation such as semantic similarity, logical inference and sentence likelihood. While we use these features and aspects as the main building blocks of NUBIA, the specific architecture and datasets can change with an attention to maintaining the necessary performance in terms of correlation with human judgment.
 
The aggregator uses the features extracted by the Transformers as well as non neural features such as reference and candidate sentence length and is trained to predict the quality of the hypothesis sentence given the reference sentence. Similar to the WMT challenge, we use past years' data to train this aggregator and test it on the test subset. 

The calibrator is the final module that caps all predictions to be between 0 and 1.

\begin{figure*}
\centering
\includegraphics[scale=0.8]{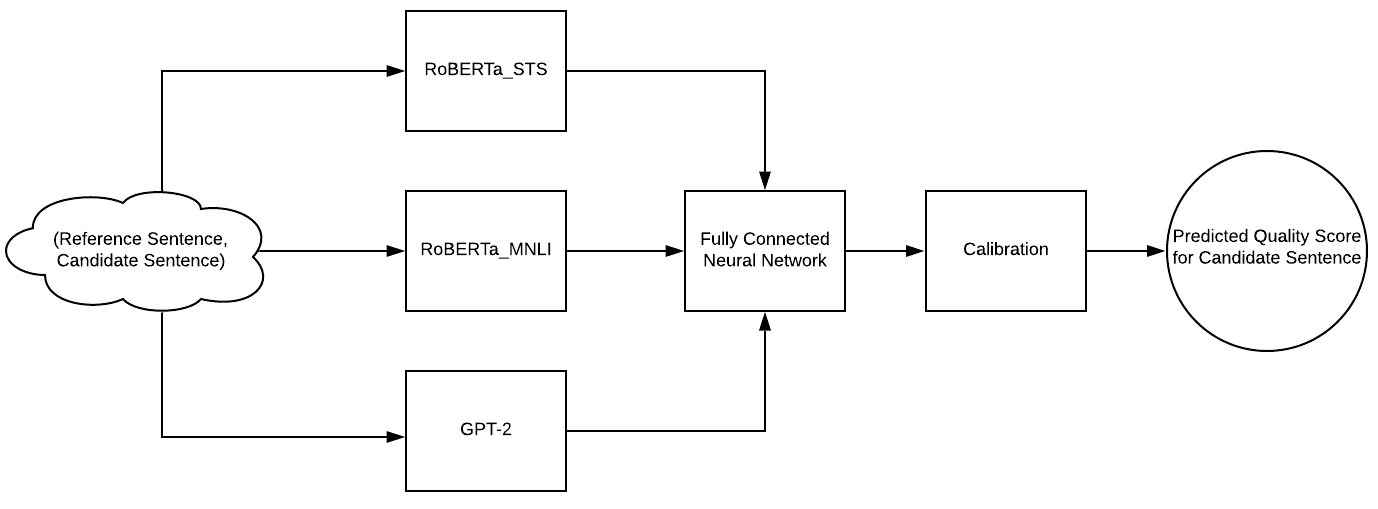}
\caption{\label{model_desing}
NUBIA steps shown with model names used in the experiments section. The three steps are Neural Feature Extraction, Aggregation and Calibration.
}
\end{figure*}

\subsection{Neural Feature Extraction }

In this section, we will describe how we broke down assessing the quality of a sentence into numerical features and also explain the thought process behind the features used. 

\subsubsection{Semantic similarity}

The first feature extracted between candidate and reference sentence is semantic similarity. To do so, we use a RoBERTa large pre-trained model \cite{liu2019roberta}, which we fine tune to predict sentence similarity (0-5 scale) on the STS-B benchmark dataset (8628 sentence pairs).

The rational between this feature is that a good candidate sentence should have high semantic similarity to the reference sentence. 

\subsubsection{Logical Inference}

The second set of features looks at the logical relationship between the reference and hypothesis sentence. The quality of the generated text depends not only on the grammar and semantics but also the core meaning and argument of the candidate sentence. A good model will output sentences that convey the same message.

To extract these features, we use a RoBERTa large pre-trained model \cite{liu2019roberta} which is then fine tuned on the MNLI challenge from the GLUE benchmark. 

The MNLI model is trained to predict a discrete score that when the location of the maximum value is taken will correspond to a discrete number with 0 meaning that the sentences are in contradiction with each other, 1 meaning that the logical relationship is undetermined/neutral and 2 meaning that the sentence are in logical agreement with each other. It outputs a distribution representing its belief over the logical relationship between the pairs of sentences it is fed as input.

For MNLI features, we take the likelihood scores over the 3 possible classes as features.

\subsubsection{Sentence Legibility}

The third set of neural features aim to capture the linguistic acceptability of the candidate sentence. 

The rationale of this feature is that we want to make sure that candidate sentences are legible and grammatically correct. 

It is a common failure mode for machine translation models to generate sentences which are close in meaning to the reference sentence but introduce uncommon syntax and grammatical errors.We currently model this by using the perplexity score of a state-of-the-art Neural Language Model: GPT-2 \cite{radford2018improving}

More precisely, given a sentence A and a sentence B, the 2 features we compute are the perplexity scores for sentence A and sentence B. Optionally, in one of the NUBIA version, we also introduce the number of words in the candidate and reference sentences. We have experimentally found that adding these features in conjunction with the perplexity scores improve correlation with human judgment.

\subsection{Aggregator}

In the section above, we defined the dimensions used to assess the quality of a candidate sentence and then showed how to turn these dimensions into features using neural networks. The aggregator module is trained to approximate a function mapping input neural features to a quality score reflecting how interchangeable the candidate sentence and the reference sentences are. 
 
The inspiration behind this model is that when human evaluators assess the quality of a candidate sentence against a reference sentence, they simultaneously pay attention to several aspects of the candidate sentence such as its semantic similarity with the reference sentence and whether it makes grammatical sense.
 
Since the relationship between these features and human judgement of quality is a priori unknown, the goal of the aggregator is to approximate it using data obtained from rigorously conducted and normalized human evaluations.

The aggregator is a regression model trained to predict human evaluation on pairs of candidate and reference sentences. In this work, we explored linear regression and feed-forward, fully connected neural network architectures.

The neural network aggregator is a fully-connected, feedforward neural network architectures with either 6 (neural features only) or 8 (neural features and number of words in candidate and reference sentences) input layers corresponding to the features extracted, 10 hidden layers and a 1 dimension output layer corresponding to the human score prediction. The activation function for the model is the hyperbolic tangent and the optimizer is ADAM \citep{kingma2014adam}. NUBIA models using 6 input features have the NUBIA-6DIM prefix while the NUBIA models using 8 input features have the NUBIA-8DIM prefix. Models using a neural network as an aggregator have the -NN suffix while those using linear regression have LREG suffix.

\subsection{Calibration}

In practice, the output of the regressors are already highly correlated with human judgement ;however, they lack two important properties. The first one is that, the regressed score comparing a reference sentence with itself is not always equal to 1. To remedy to this, we normalize the scores given to a candidate sentence by the score given by the regressor of the candidate sentence with itself. The second missing property is that the raw regression scores are not strictly bounded to be between 0 and 1. To ensure they are, we cap the output of the regressors to have a value between 0 and 1.

\section{Experiments}

To assess NUBIA, we used both direct assessment and segment-level relative ranking from different WMT metrics shared tasks \cite{ufal2017results} \cite{graham2015accurate} as well as tasks from image captioning. We did not conduct experiments in the domain of machine summarization because while datasets exists, there are no labeled datasets paired with human evaluations.

In the Direct Assessment task, candidate and reference translations are given for several language pairs and for each candidate translation, 15 human evaluators assign a quality score between 0 and 100. The final human score is taken as the average of the 15 human assessments.The performance of metrics is assessed using pearson correlation with human judgement. For this task, we used the 2017 dataset because, unlike the WMT 2018 and WMT 2019 dataset, each sentence has been scored by at least 15 human evaluators \citet{ma-etal-2018-results}. 
For relative ranking, WMT 2018 and WMT 2019 still use direct human assessments but since there is not at least 15 annotators per sentence, the direct assessment correlation task is converted into relative ranking task. More specifically, for a given reference sentences, up to 5 machine translation systems generate candidate translations. These candidate sentences are rated by human annotators on a discrete 0-25-50-75-100 points scale. After averaging the human annotations, if the gap between two candidate translation is higher than 25 points, one translation is considered to be better than the other. When the gap between two candidate sentences is higher than 25 points, the sentence pairs are not included in the segment-level evaluation \citet{ma-etal-2018-results}. In that setting, metrics are scored on their ability to preserve the human ranking using the Kendall's Tau correlation coefficient.
\begin{figure*}[!tbp]
  \centering
  \begin{minipage}[b]{0.3\textwidth}
    \includegraphics[scale = 0.14]{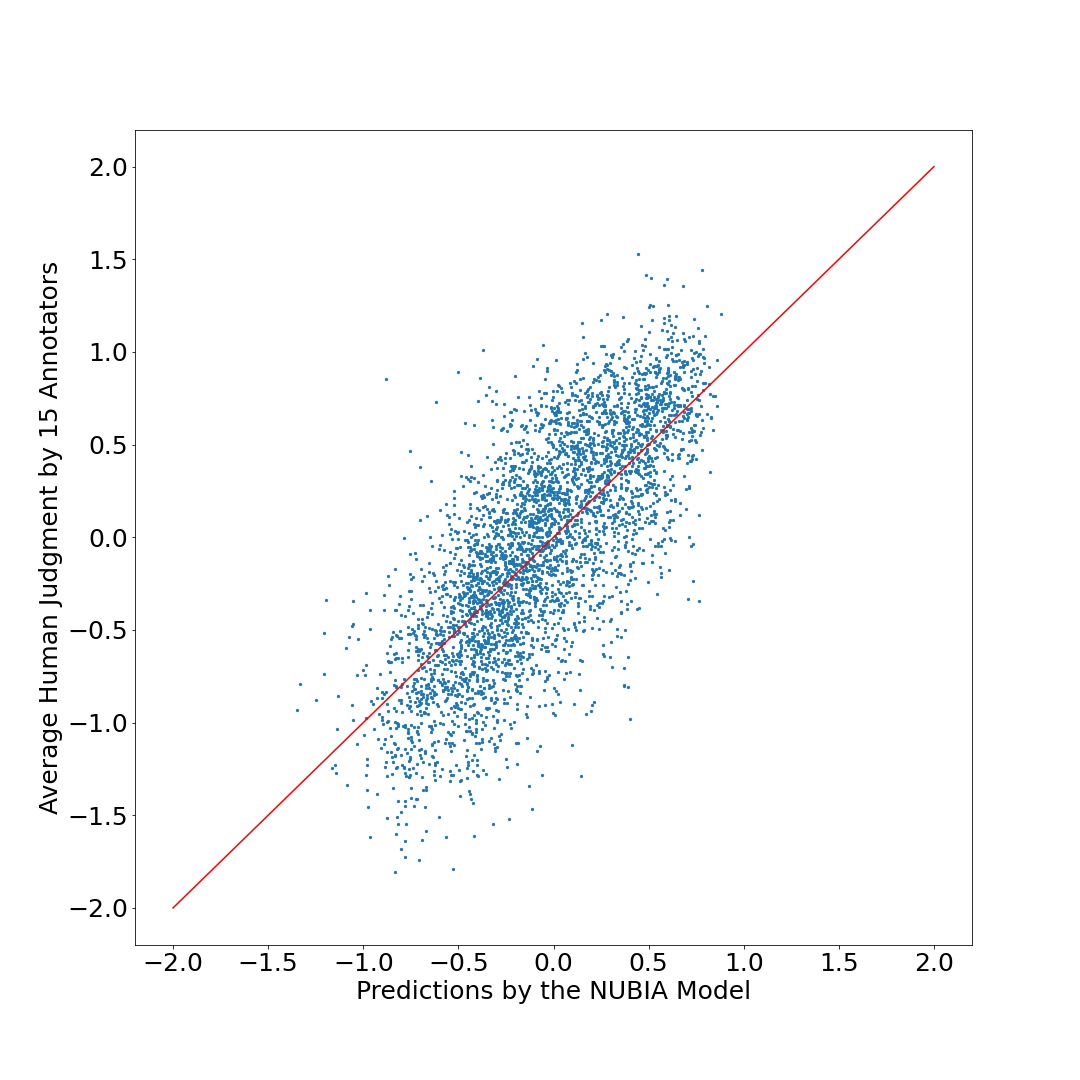}
    \subcaption{NUBIA-NN}
  \end{minipage}
  \hfill
  \begin{minipage}[b]{0.3\textwidth}
    \includegraphics[scale = 0.14]{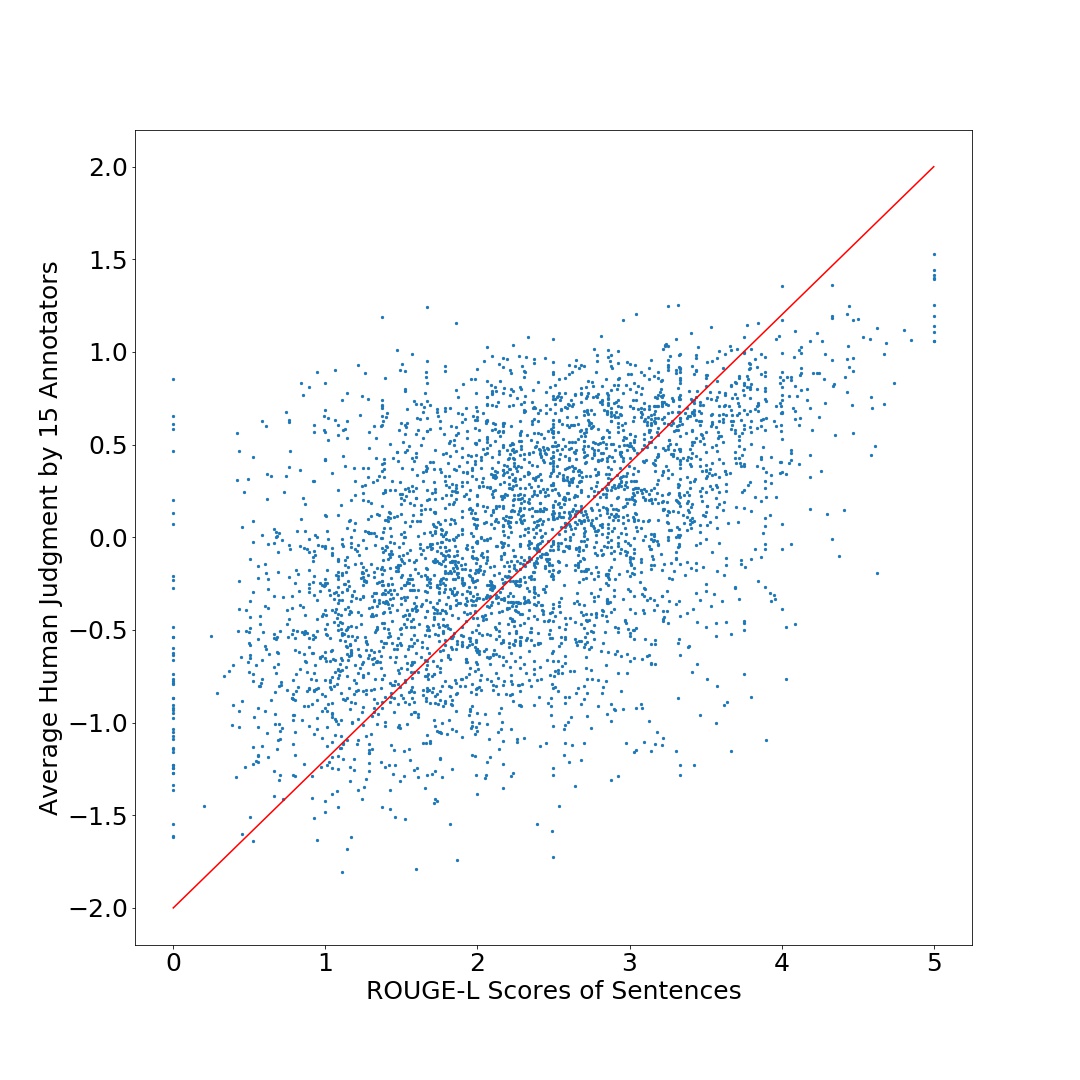}
    \subcaption{ROUGE-L}
  \end{minipage}
  \hfill
  \begin{minipage}[b]{0.3\textwidth}
    \includegraphics[scale = 0.14]{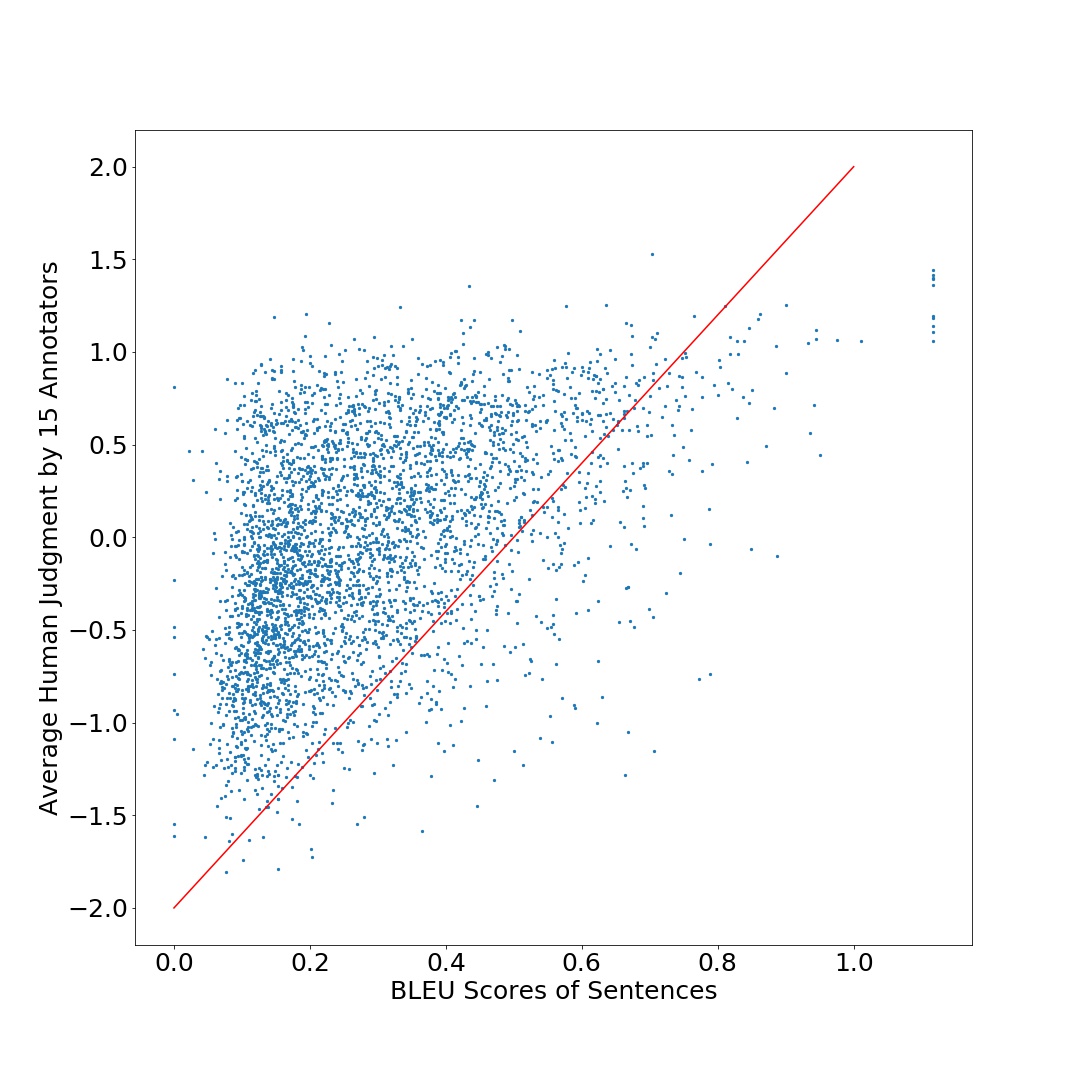}
    \subcaption{BLEU}
  \end{minipage}
  \caption{Score and label graphs of NUBIA, ROUGE-L and BLEU for the entire WMT-2017 segment level sets.}
  \label{visuals}
\end{figure*}

\subsection{Model Training}

For our training set, we use the WMT 2015, WMT 2016, WMT 2017 and WMT 2018 datasets in different settings. All datasets are used for testing in future years. In these datasets, we only picked translations where the target language is English. This was done because the language models we used and their underlying word embeddings are trained on English sentences. 

This gives us 2000 sentence pairs for WMT2015, 3360 pairs for WMT2016, 3920 sentence pairs for WMT2017. The described datasets give us a training set of 5360 sentence pairs for training on WMT2017, 9280 sentences for training on WMT2018 and WMT2019. In practice we found no improvement by adding sentences from WMT2018 to train the aggregator which is why we stick with WMT 2015 through 2017 to test on both WMT 2018 and WMT 2019.

\subsection{Model Testing}
For testing, we run experiments on machine translation and image captioning. 

\subsubsection{Machine Translation}
For our test set, we used English sentence pairs on the WMT2017, WMT2018 and WMT2019 datasets. We extracted features using our feature extractor and fed them to our trained aggregators to predict the human quality score. Our test set has 3920, 207576, and 281009 sentence pairs. 

For the WMT2017 task the testing method is Pearson correlation. This method is replaced in the WMT2018 and WMT2019 challenges with relative ranking where for each reference sentence a number of hypothesis sentences are scored and if the a given hypothesis scores better than 25 points out of a 100 than another hypothesis sentence it is marked as a better sentence. Metrics are compared with a Kendall's Tau formulation on how well their scores correlate with human scores. 

\subsubsection{Image Captioning}
For image captioning we followed SPICE and used the Flickr 8K dataset. This dataset consists of 8092 images annotated with 5 gold standard captions generated by humans. The dataset also has a human evaluated part where for each image, a new caption is selected from the entire dataset and scored by three expert judges between 1("the selected caption is unrelated to the image") and 4("the selected caption describes the image without any error.). This part has 5822 human evaluated image caption pairs where each image also has 5 reference gold standard captions. NUBIA is compared with Kendall's Tau on how well it correlates with the average of the three judges' scores as labels. The aggregators for the NUBIA models used in the image captioning experiments are not specifically fine tuned for the task and consist of the Neural Feature Extractors described above along with an aggregator trained on the WMT2015, WMT2016 and WMT2017 dataset.
\section{Results}

\begin{table*}[h]
\centering
\begin{tabular}{ll}
\hline
\textbf{} & \textbf{Flickr 8K} \\
\hline
BLEU-1* &0.32\\
BLEU-4 &0.33\\
ROUGE-L* & 0.32\\
METEOR* &0.42\\
CIDEr* &0.44\\
SPICE* &0.45 \\
\textbf{NUBIA-6DIM-NN} & \textbf{0.47} \\

\hline
\end{tabular}
\caption{\label{sec:image_captioning}
Kendall's Tau Correlation with human judgment on Flickr 8K dataset. The scores marked with * are taken directly from the original SPICE paper. The BLEU-4 score in the original paper was 0.14 but the experiment was repeated with a smoothed function and the new result is reported.
}
\end{table*}

\begin{table*}[h]
\centering
\begin{tabular}{lllllllll}
\hline
\textbf{} & \textbf{cs-en} & \textbf{de-en}& \textbf{fi-en}& \textbf{lv-en}& \textbf{ru-en}& \textbf{tr-en}& \textbf{zh-en}& \textbf{AVG}\\
\hline
\begin{tabular}{@{}c@{}}Human Evaluation \\ \textit{n} \\Correlation \end{tabular} & \begin{tabular}{@{}c@{}}DA \\ 560 \\ $|r|$ \end{tabular} & \begin{tabular}{@{}c@{}}DA \\ 560 \\ $|r|$ \end{tabular} & \begin{tabular}{@{}c@{}}DA \\ 560 \\ $|r|$ \end{tabular} & \begin{tabular}{@{}c@{}}DA \\ 560 \\ $|r|$ \end{tabular} & \begin{tabular}{@{}c@{}}DA \\ 560 \\ $|r|$ \end{tabular} & \begin{tabular}{@{}c@{}}DA \\ 560 \\ $|r|$ \end{tabular} & \begin{tabular}{@{}c@{}}DA \\ 560 \\ $|r|$ \end{tabular} & \begin{tabular}{@{}c@{}}DA \\ 3920 \\ $|r|$ \end{tabular} \\
\hline
BLEU &0.432& 0.425& 0.577& 0.415& 0.479& 0.548& 0.515 &0.484\\
ROUGE-L &0.482& 0.492 & 0.623 & 0.465 & 0.480 & 0.593 & 0.569 &0.529\\
BLEND & 0.594 & 0.571 & 0.733 & 0.577 & 0.622& 0.671& 0.661 &0.632\\
MEANT2.0&0.578&0.565&0.687&0.586&0.607&0.596&0.639&0.608\\
RUSE &0.614&  0.637&0.756 & 0.705 & 0.680 &0.704 & 0.677&0.681\\
BERTscore &0.714  & \textbf{0.740}  & 0.835  & 0.774 & \textbf{0.773}  & 0.776 &  \textbf{0.767}&0.768\\
NUBIA-8DIM-LReg &0.738 & 0.732& 0.828& \textbf{0.783}& 0.731& 0.782& \textbf{0.768}&0.766\\
NUBIA-8DIM-NN &\textbf{0.753} & \textbf{0.738}  & \textbf{0.854}  & \textbf{0.785} & 0.755  & \textbf{0.804} & 0.750 &\textbf{0.777}\\

\hline
\end{tabular}
\caption{\label{sec:wmt17}
Absolute  Pearson  correlations  with  segment-level  human  judgments  on  WMT17  to-English translations.  Correlations of metrics not significantly outperformed by any other for that language pair are highlighted in bold.
}
\end{table*}

\begin{table*}[h]
\centering
\begin{tabular}{lllllllll}
\hline
\textbf{} & \textbf{cs-en} & \textbf{de-en}& \textbf{et-en}& \textbf{fi-en}& \textbf{ru-en}& \textbf{tr-en}& \textbf{zh-en}& \textbf{AVG}\\
\hline
\begin{tabular}{@{}c@{}}Human Evaluation \\ \textit{n} \\Correlation \end{tabular} & \begin{tabular}{@{}c@{}}DA \\ 560 \\ $|r|$ \end{tabular} & \begin{tabular}{@{}c@{}}DA \\ 560 \\ $|r|$ \end{tabular} & \begin{tabular}{@{}c@{}}DA \\ 560 \\ $|r|$ \end{tabular} & \begin{tabular}{@{}c@{}}DA \\ 560 \\ $|r|$ \end{tabular} & \begin{tabular}{@{}c@{}}DA \\ 560 \\ $|r|$ \end{tabular} & \begin{tabular}{@{}c@{}}DA \\ 560 \\ $|r|$ \end{tabular} & \begin{tabular}{@{}c@{}}DA \\ 560 \\ $|r|$ \end{tabular} & \begin{tabular}{@{}c@{}}DA \\ 3920 \\ $|r|$ \end{tabular} \\
\hline
BLEU &0.268& 0.458& 0.311& 0.206& 0.259& 0.178&0.21 &0.27\\
ROUGE-L &0.28& 0.473 & 0.324 & 0.208 & 0.275 & 0.193 & 0.211 &0.281\\
YiSi-1-srl (WMT18 version) &0.317 & 0.483 & 0.345 & 0.237 & 0.306& 0.233& 0.209 & 0.304\\
RUSE&0.3478&0.498&0.368&0.273&0.311&0.259&0.218&0.325\\
YiSi-1-srl &0.396&  0.543&0.39 & 0.303 & 0.351 &0.297 & 0.253&0.362\\
Yisi-1 &0.391  & 0.544  & 0.397  & 0.299 & 0.352  & \textbf{0.301} &  0.254&0.363\\
BERTScore &\textbf{0.408} & \textbf{0.55}& 0.395& 0.293& 0.346& 0.296& 0.26&0.364\\
NUBIA-8DIM-NN &0.382 & 0.544  & \textbf{0.406}  & \textbf{0.324} & \textbf{0.358}  & 0.298 & \textbf{0.263} &\textbf{0.368}\\

\hline
\end{tabular}
\caption{\label{sec:wmt18}
Kendall's Tau correlation  with  segment-level  human  judgments  on  WMT18  to-English translations.  Correlations of metrics not significantly outperformed by any other for that language pair are highlighted in bold.
}
\end{table*}

\begin{table*}[h]
\centering
\begin{tabular}{lllllllll}
\hline
\textbf{} & \textbf{de-en} & \textbf{fi-en}& \textbf{gu-en}& \textbf{kk-en}& \textbf{lt-en}& \textbf{ru-en}& \textbf{zh-en}& \textbf{AVG}\\
\hline
\begin{tabular}{@{}c@{}}Human Evaluation \\ \textit{n} \\Correlation \end{tabular} & \begin{tabular}{@{}c@{}}DA \\ 560 \\ $|r|$ \end{tabular} & \begin{tabular}{@{}c@{}}DA \\ 560 \\ $|r|$ \end{tabular} & \begin{tabular}{@{}c@{}}DA \\ 560 \\ $|r|$ \end{tabular} & \begin{tabular}{@{}c@{}}DA \\ 560 \\ $|r|$ \end{tabular} & \begin{tabular}{@{}c@{}}DA \\ 560 \\ $|r|$ \end{tabular} & \begin{tabular}{@{}c@{}}DA \\ 560 \\ $|r|$ \end{tabular} & \begin{tabular}{@{}c@{}}DA \\ 560 \\ $|r|$ \end{tabular} & \begin{tabular}{@{}c@{}}DA \\ 3920 \\ $|r|$ \end{tabular} \\
\hline
BLEU &0.173& 0.264& 0.207& 0.389& 0.280& 0.166& 0.349 &0.261\\
ROUGE-L &0.169& 0.268 & 0.198 & 0.394 & 0.294 &0.171 & 0.348 &0.263\\
ESIM & 0.167 & 0.337 &0.303 & 0.435 & 0.359& 0.201& 0.396 &0.314\\
NUBIA-8DIM-NN&\textbf{0.238}&\textbf{0.349}&0.260&0.411&0.374&\textbf{0.223}&0.409&0.323\\
YISI &0.199& 0.346&0.306 & \textbf{0.442} & \textbf{0.380} &0.222 & 0.431&0.332\\
BERTscore &0.230  & 0.345  & \textbf{0.320}  & 0.432 & \textbf{0.381}  & \textbf{0.223} &  \textbf{0.444}&\textbf{0.339}\\
\hline
\end{tabular}
\caption{\label{sec:wmt19}
Kendall's Tau correlation  with  segment-level  human  judgments  on  WMT19  to-English translations.  Correlations of metrics not significantly outperformed by any other for that language pair are highlighted in bold.
}
\end{table*}

In Table~\ref{sec:wmt17}, we report our results on the test set. We compare our methods with methods developed for the WMT2017 challenge and models like RUSE and BERTScore which are currently the best performing methods. Although many methods have been proposed throughout the years in the WMT metrics challenge, the current methods used to this day to assess performance of summarization and translation models are still BLEU and ROUGE score. For ROUGE, we use ROUGE-L scores because it is the formulation of ROUGE correlated the most with human judgements on WMT 2017.

In Table-\ref{sec:wmt18}, we report the results for the relative ranking test of MWT2018. Here we see that NUBIA achieves state of the art results outperforming all metrics in 5 out of 8 language pairs. In Table-\ref{sec:wmt19}, we have the results for the WMT2019 challenge. Here we observe that NUBIA performs comparably with other methods. 

We report the results of the image captioning experiments in Table-\ref{sec:image_captioning}. Here we observe that NUBIA outperforms all existing methods and achieves state of the art correlation with human judgment of caption quality.  

\subsection{Ablation Study}

To judge the importance of the features we have picked, we ran an ablation study where we trained a NUBIA model with only a subset of the features and report correlation results on the WMT17 dataset. The most crucial feature is the RoBERTa semantic similarity score. As suspected, other elements beyond semantic similarity also seem to be factored into prediction of translation quality as evidenced by the performance boost obtained after computing the GPT-2 features and MNLI features.

\begin{table*}
\centering
\begin{tabular}{lllllllll}
\hline
\textbf{} & \textbf{cs-en} & \textbf{de-en}& \textbf{fi-en}& \textbf{lv-en}& \textbf{ru-en}& \textbf{tr-en}& \textbf{zh-en}&\textbf{AVG}\\
\hline
\begin{tabular}{@{}c@{}}Human Evaluation \\ \textit{n} \\Correlation \end{tabular} & \begin{tabular}{@{}c@{}}DA \\ 560 \\ $|r|$ \end{tabular} & \begin{tabular}{@{}c@{}}DA \\ 560 \\ $|r|$ \end{tabular} & \begin{tabular}{@{}c@{}}DA \\ 560 \\ $|r|$ \end{tabular} & \begin{tabular}{@{}c@{}}DA \\ 560 \\ $|r|$ \end{tabular} & \begin{tabular}{@{}c@{}}DA \\ 560 \\ $|r|$ \end{tabular} & \begin{tabular}{@{}c@{}}DA \\ 560 \\ $|r|$ \end{tabular} & \begin{tabular}{@{}c@{}}DA \\ 560 \\ $|r|$ \end{tabular} & \begin{tabular}{@{}c@{}}DA \\ 3920 \\ $|r|$ \end{tabular} \\
\hline

NUBIA-NN,LI &
0.620&
0.539&
0.693&
0.647&
0.603&
0.692&
0.571&0.623\\
NUBIA-NN,SI &
0.412&
0.451&
0.624&
0.571&
0.447&
0.437&
0.410&0.478\\
NUBIA-NN,SS &
0.678&
0.686&
0.790&
0.740&
0.694&
0.766&
0.708&0.723\\
NUBIA-NN,LI+SI &
0.643&
0.621&
0.775&
0.722&
0.646&
0.681&
0.624 &0.673\\
NUBIA-NN,SS+LI &
0.696&
0.699&
0.804&
0.758&
0.708&
0.784&
0.723&0.738
\\
NUBIA-NN,SS+SI&
0.727&
\textbf{0.729}&
\textbf{0.842}&
\textbf{0.785}&
0.726&
0.790&
0.755&0.764\\
NUBIA-NN,SS+LI+SI &\textbf{0.753} & \textbf{0.738}  & \textbf{0.854}  & \textbf{0.784} & \textbf{0.755}  & \textbf{0.804} & \textbf{0.750}&\textbf{0.777}\\

\hline
\end{tabular}
\caption{\label{sec:ablation}
Ablation study results for NUBIA-NN on WMT 2017 Direct Assessment task. SS=Semantic Similarity, LI=Linguistic Inference, SI=Sentence Intelligibility.
}
\end{table*}

\subsection{Error Analysis}
Figure \ref{visuals} sheds more light on the behavior of BLEU and ROUGE, two of the most common evaluation metrics and NUBIA-NN. This analysis unveils important properties of these metrics and helps better understand their strengths and weaknesses.

If we start with (c) we can see that BLEU correlates better with Human Judgment in the bottom left (bad hypothesis area). Essentially, if a human is likely to give a bad score to a sentence, BLEU is unlikely to overscore. But if a person is going to give a high score, BLEU is equally likely to give any score, maybe even more likely to penalize the sentence. This effectively inhibits the desired behaviour in language generation. 

This behaviour is of course not all unwanted. In the early days BLEU could be seen as a harsh and firm critic, but now we need more robust evaluators of good candidate sentences. 

While the behaviour of ROUGE is much more balanced, it is still prone to underscoring and over-scoring.

When we look at NUBIA-NN, we see a general trend followed along the data, as expected given the high correlation score. The only interesting action is the over scoring of low human score sentences. While this behaviour is not likely to cause a large margin of error or a big opportunity for exploitation, the nature of the error can be analyzed to further improve NUBIA.

\section{Conclusion}

In this work, we introduced NUBIA: a methodology to build automatic  evaluation  metrics  for  text  generation  using machine  learning  models as core components. This methodology achieves state-of-the-art results across evaluation of machine translation and image captioning strongly building on the successes of recent NLP architectures such as RoBERTa and GPT-2. These strong results suggest that using a neural networks to extract features and combine them will be a key component of building future automatic scoring metrics for text generation with the promise of unifying evaluation of image caption, machine translation and potentially other text generation tasks.

\section{Discussion and future work}

Learned text generation evaluation metrics have enormous promise to change how text generation models are assessed. Future work can further probe which other text generation tasks NUBIA models are strong candidates to assess.

NUBIA can be improved through three axis. The first axis of improvement is through the efforts of the wider NLP community at creating models achieving strong results on the NLU benchmarks like GLUE. The second axis is through the addition of better features capturing aspects of human quality assessment. Two candidate features are the linguistic acceptability which can be obtained by using models trained on the CoLA challenge and a coherence score for long text generations. The third axis is through better aggregator design. 

Learning how to specify the NUBIA architectures and standardizing nomenclature will be crucial to ensure adoption, reproducibility and fair comparison of models scored using such automatic metrics. An exhaustive solution can be to describe the individual feature extractor. This description should not only include architectures but also training data and fine tuning data. Similarly, aggregators should also be described through their architectures along with the training corpus. Model cards \citep{mitchell2019model} and better metric assessments going beyond correlation with human judgement \citep{kane2019towards} \citep{boag2016mutt} will be key components of improved model reporting.

Another area of current limitation is the language. Existing NUBIA models only work for English sentence pairs though the procedure to generate and assess such metrics in other languages is likely to be similar. 

Other areas of vulnerability can also include biased training data leading to underscoring or over scoring valid translations. Data statements \citep{bender2018data} or data sheets for datasets \citep{gebru2018datasheets} on individual components and the NUBIA metric can help design more transparent/trustworthy system.

Finally, understanding how such models can be adversarially attacked is also an open research question. 

\bibliography{anthology,acl2020}
\bibliographystyle{acl_natbib}

\end{document}